\documentclass[runningheads]{llncs}
\usepackage{graphicx}
\usepackage{microtype}
\usepackage{amsmath,amssymb,amsfonts}
\usepackage{algorithmic}
\usepackage{textcomp}
\usepackage{xcolor}
  \usepackage{hyperref}
\usepackage{cleveref}
\usepackage{fixltx2e}
 \usepackage{multirow}
  \usepackage{makecell}
  \usepackage{times}
\usepackage{latexsym}
\usepackage{mathtools}
\usepackage{multirow}
\usepackage{makecell}
\usepackage{float}
\usepackage{lipsum}
\usepackage{amsmath}
\usepackage{cleveref}
\restylefloat{table}    
\usepackage{arydshln}
\usepackage{adjustbox}
\usepackage{subfig}
\usepackage{cite}
\begin{document}
\def\methodname/{\mbox{XCoref}}

\title{XCoref: Cross-document Coreference \\ Resolution in the Wild}
\titlerunning{XCoref: Cross-document Coreference Resolution in the Wild}
%
\author{Anastasia Zhukova\inst{1}\orcidID{0000-0001-9084-2890} \and
Felix Hamborg\inst{2,4}\orcidID{0000-0003-2444-8056} \and
Karsten Donnay\inst{3,4}\orcidID{0000-0002-9080-6539} \and 
Bela Gipp\inst{1,4}\orcidID{0000-0001-6522-3019}}
\authorrunning{Zhukova et al.}
%
\institute{Data and Knowledge Engineering, University of Wuppertal, Germany
\email{\{last\}@uni-wuppertal.de}
 \and
Dept. of Computer Science, University of Konstanz, Germany 
\email{felix.hamborg@uni-konstanz.de}
\and 
Dept. of Political Science, University of Zurich, Switzerland 
\email{donnay@ipz.uzh.ch}
\and 
Heidelberg Academy of Sciences and Humanities, Germany
}

\maketitle              
\begin{abstract}
Datasets and methods for cross-document coreference resolution (CDCR) focus on events or entities with strict coreference relations. They lack, however, annotating and resolving coreference mentions with more abstract or loose relations that may occur when news articles report about controversial and polarized events. Bridging and loose coreference relations trigger associations that may lead to exposing news readers to bias by word choice and labeling. For example, coreferential mentions of ``direct talks between U.S. President Donald Trump and Kim'' such as ``an extraordinary meeting following months of heated rhetoric'' or ``great chance to solve a world problem'' form a more positive perception of this event. A step towards bringing awareness of bias by word choice and labeling is the reliable resolution of coreferences with high lexical diversity. We propose an unsupervised method named \textit{\methodname/}, which is a CDCR method that capably resolves not only previously prevalent entities, such as persons, e.g., ``Donald Trump,'' but also abstractly defined concepts, such as groups of persons, ``caravan of immigrants,'' events and actions, e.g., ``marching to the U.S. border.'' In an extensive evaluation, we compare the proposed \methodname/ to a state-of-the-art CDCR method and a previous method TCA that resolves such complex coreference relations and find that \methodname/ outperforms these methods. Outperforming an established CDCR model shows that the new CDCR models need to be evaluated on semantically complex mentions with more loose coreference relations to indicate their applicability of models to resolve mentions in the ``wild'' of political news articles.

\end{abstract}

\section{Introduction}

Coreference resolution (CR) is a set of techniques that aim to resolve mentions of entities, often in a single text document. CR is employed as an essential analysis component in a broad spectrum of use cases, e.g., to identify potential targets in sentiment analysis or as a part of discourse interpretation. While CR focuses on single documents, \textit{cross-document coreference resolution} (CDCR) resolves concept mentions across a set of multiple documents. Compared to CR, CDCR is a less-researched task, although is more challenging due to a larger search space and much required to facilitate content understanding of multiple articles. Further, many use cases require CDCR of more varying concepts than just entities typically resolved within CR, e.g., events or abstract entities. 

Although the CDCR research has been gaining attention, the annotation schemes and corresponding datasets have been infrequently exploring a mix of identity and loose coreference relations and lexical diversity of the annotated chains of mentions. We explore CDCR in a particularly challenging use case, i.e., to identify bias by word choice and labeling. Such bias occurs due to strong variance in the words and loose coreference relations that yield possibly biased interpretations of events or entities.  For example, coreferential mentions of ``direct talks between U.S. President Donald Trump and Kim'' such as ``an extraordinary meeting following months of heated rhetoric'' or ``great chance to solve a world problem'' form a more positive perception of this event. 

Resolution of identity relations, i.e., coreference resolution, and resolution of more loose relations, i.e., bridging, are typically split into two separate tasks \cite{kobayashi-ng-2020-bridging}. Resolution and evaluation of entity and event mentions of the two types of relation remains a research gap in general CDCR research. Hamborg et al. \cite{Hamborg2019a} first explored CDCR in a particularly challenging use case, i.e., to identify bias by word choice and labeling in news articles. Their proposed approach called TCA resolved mentions with strong lexical diversity. 

In this paper, first, we revisit TCA and propose \methodname/, an unsupervised sieve-based method that resolves jointly mentions of strict and loose identity relations into coreferential chains\footnote{For the peer review, an anonymous link to the code:  \url{https://drive.google.com/file/d/18vfxdSwXE9fsjEq6IYi1j9cDTJNDoOXX/}}. Methods of such design have been successfully used to resolve mentions of identity relation in events and entities \cite{ ijcai2018-773, lee2011stanford}, and bridging relations in entities \cite{Hou2018a}. 

Second, we conduct an extensive evaluation where we compare \methodname/ with TCA and one of the state-of-the-art method for CDCR \cite{barhom-etal-2019-revisiting}. We evaluate the annotated mentions of varying coreference strength jointly as one task of set identification and calculate the results with the standard CoNLL metrics in (CD)CR \cite{pradhan-etal-2012-conll}. We discuss a direction of the CDCR evaluation to address the complexity of coreferential chains annotated on diverse political news articles, i.e., in the ``wild.'' 

\section{Related work}
\label{sec:rel}

Coreference resolution (CR) and cross-document coreference resolution (CDCR) are tasks that aim to resolve coreferential mentions in one or multiple documents, respectively \cite{Singh_2019}. (CD)CR approaches tend to depend on the annotation schemes of the CDCR datasets that specify the definition of mentions and coreferential relations \cite{bugert2021generalizing}.

Most (CD)CR datasets contain only strict identity relations, e.g., TAC KBP \cite{mitamura2017events, mitamura2015overview}, ACE \cite{linguistic2008ace, linguistic2005ace}, MEANTIME \cite{minard-etal-2016-meantime}, OntoNotes \cite{weischedel2011ontonotes}, ECB+ \cite{bejan2010unsupervised, cybulska-vossen-2014-using}. Less commonly used (CD)CR datasets explore relations beyond strict identity. For example, NiDENT \cite{recasens-etal-2012-annotating} is a CDCR dataset of entities-only mentions that was created by reannotating NP4E. NiDENT explores coreferential mentions of more loose coreference relations coined near-identity that among all included metonymy, e.g., ``White House'' to refer to the US government, and meronymy, e.g., ``US president'' being a part of the US government and representing it. Reacher Event Description (RED), a dataset for CR, contains also more loose coreference relations among events \cite{ogorman-etal-2016-richer}. 

Mentions coreferential with more loose relations are harder to annotate and automatically resolve than mentions with identity relations \cite{recasens2010typology}. Bridging relations occur when a connection between mentions is implied but is not strict, e.g., a ``part-of'' relation. Bridging relations, unlike identity relations, form a link between nouns that do not match in grammatical constraints, e.g., gender and number agreement, and allow linking noun and verb mentions, thus, constructing abstract entities \cite{kobayashi-ng-2020-bridging}. The existing datasets for identification of bridging relations, e.g., ISNotes \cite{Hou2018a}, BASHI \cite{rosiger-2018-bashi}, ARRAU \cite{poesio-artstein-2008-anaphoric}, annotate the relations only of noun phrases on a single-document level and solve the problem as antecedent identification problem rather than identification of a set of coreferential anaphora \cite{Hou2018a}. Definition identification in DEFT dataset \cite{spala-etal-2019-deft} focuses on annotating mentions that are linked with ``definition-like'' verb phrases (e.g., means, is, defines, etc.) but does not address linking the antecedents and definitions into the coreferential chains.

To our knowledge, only one dataset contains annotations of coreferential mentions with varying strength of coreferential relations. NewsWCL50 \cite{Hamborg2019a} contains annotations of concepts based on a minimum number of (coreferential) mentions across a set of news articles reporting on the same event. The dataset contains diverse concept types, such as actors, entities, events, geo-political entities (GPEs), and also more complex types, such as actions or abstract entities. The dataset argues that in political news articles more loose coreferential relations form links and associations to phrases that could bear bias by word choice and labeling, e.g., ``DACA recipients'' -- ``undocumented immigrants who came to the U.S. as children'' -- ``illegal aliens'' -- ``innocent kids.''


There are two approaches for event CDCR, easy-first and mention-pair \cite{ijcai2018-773}. Usually, easy-first approaches are unsupervised, whereas mention-pair are supervised. Most methods employ the easy-first approach and sequentially execute so-called sieves. Each sieve resolves mentions concerning specific characteristics. Earlier sieves target simple and generally reliable properties, such as heads of phrases. Later sieves address more difficult or specialized cases and use special techniques, such as pair-wise scoring of the pre-identified concepts with binary classifiers, e.g., SVM \cite{lee-etal-2012-joint, lu-ng-2016-event, Hamborg2019a, nlpa2018}. Alternatively, a mention-pair approach uses a neural model trained to score the likelihood of a pair of the event- or entity-mentions to refer to the same semantic concept. The features to represent mentions are spans of text, contexts, and semantic dependencies \cite{barhom-etal-2019-revisiting}. 

Most CDCR methods focus on only events and resolve entities---if at all---as subordinate attributes of the events \cite{Keshtkaran2017, ijcai2018-773, lu-ng-2016-event, barhom-etal-2019-revisiting, Cattan2021, cybulska-vossen-2015-translating, kenyon-dean-etal-2018-resolving}. These a few CDCR methods resolve chains of both event and entity mentions with strict identity and there is only one method the resolve concepts with more loose the identity and coreference relations \cite{Hamborg2019a}. The method named TCA \cite{Hamborg2019a}, resolves mentions of varying identity levels, i.e., include strict and near-identity. Target Concept Analysis (TCA) is a sieve-based method that resolves concepts that represent entities, events, and aggregating categories, e.g., categories that include mentions referring to both a country and its governmental institutions to which they belong annotated with a name of a country \cite{Hamborg2019a}. Other sieve-based feature-engineering methods for mention resolution were successfully used for identification of entities and events with strict identity \cite{nlpa2018, ijcai2018-773, lee2011stanford}, and resolution of mentions with bridging relations \cite{hou-etal-2013-global, Hou2018a}. 

In conclusion, prior methods for CDCR suffer from at least one of two shortcomings, i.e., they (1) only resolve mentions interlinked with identity relations or (2) focus on event-driven narrowly defined coreferential mentions. The contributions of this paper are two-fold: first, we revisit the methodology of TCA because it is the only method that addresses the resolution of mentions with various identity relations \cite{Hamborg2019a}. Then, we propose \methodname/, an unsupervised method that jointly resolves mentions with strict and loose anaphoric relations. Second, we evaluate the approach on a CDCR dataset with coreferential chains with varying identity relations, i.e., NewsWCL50, and compare the results to the previously proposed CDCR methods for these datasets using metrics established in the literature on (CD)CR, i.e., B3, CEAF\_e, and MUC \cite{moosavi-strube-2016-coreference}.

\section{Methodology: \methodname/}

\begin{figure}[h]
\centering
\includegraphics[width=1.0\textwidth]{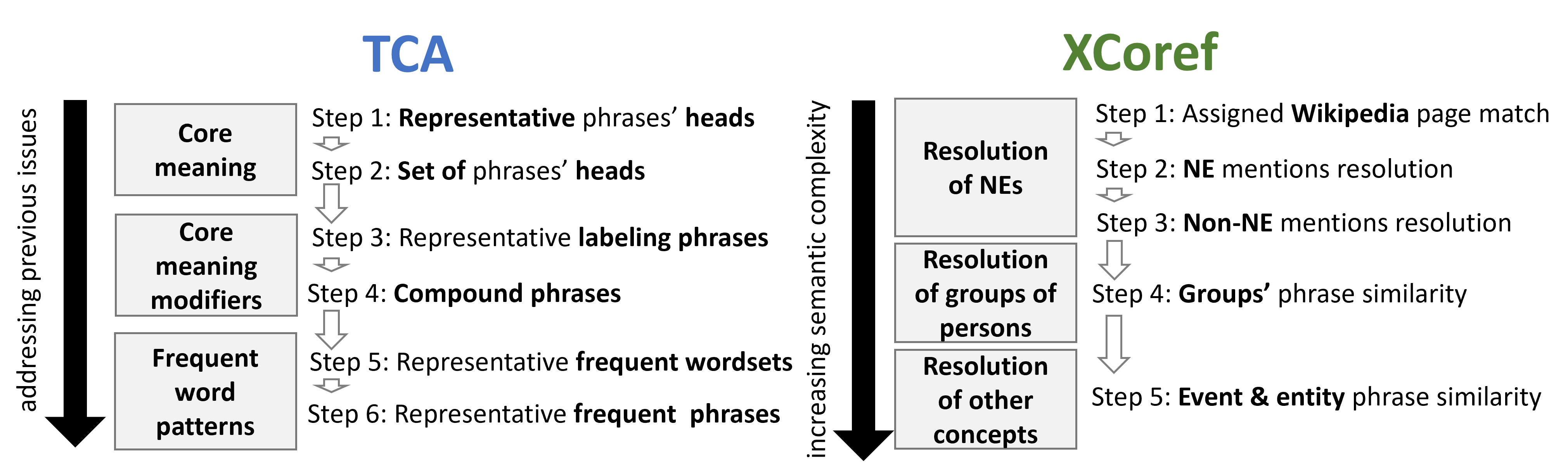}
\caption{Comparison of TCA to \methodname/: TCA resolves mentions by addressing previous issues whereas \methodname/ first resolves coreferential chains of NEs with identity relations (S1-S3) and then resolve those chains with more loose coreferential relations (S4 and S5).}
\label{img:comp}
\end{figure}

\methodname/ revisits Target Concept Analysis (TCA) proposed by Hamborg et al. \cite{Hamborg2019a}. \methodname/ consists of five sieves (see \Cref{img:comp}) and applies the ``easy-first'' principle, i.e., it first resolves mentions that belong to named entities (NEs), such as person, organization, and country, and are coreferential with identity relation \cite{Hou2018a}. Afterward, the method addresses chains coreferential with mixed identity and bridging relations, i.e., groups of persons, events, and abstract entities.  TCA is a sieve-based method to resolve coreferential chains, by addressing issues of the previously unresolved mentions. In contrast to TCA, \methodname/ resolves mentions of specific concept types in each sieve and analyzes combinations of phrases' modifiers to resolve mentions with varying coreference relations.  

\subsection{Preprocessing}
Since each sieve of \methodname/ resolves specific concept types, we first need to identify these types. We distinguish nine types: person, group, country, misc, and their NE- and non-NE variations \cite{Hamborg2019a}. Further, each sieve $S_i$ uses a comparison matrix $cm_i$, i.e., a manually created cross-type matrix, to determine which types $x$ and $y$ to compare for potential merging ($cm_{i,x,y}\geq 1$ allows comparison, $0$ not). 

To yield reliable concept type determination, we propose a type scoring based on WordNet sense ranking \cite{fellbaum2005}. For a given concept, Hamborg et al. \cite{Hamborg2019a} proposed counting frequencies of WordNet senses assigned to all concept's heads', e.g., ``president'' is four times ``noun.person.'' We propose an improvement to the counting by weighting the senses according to their rank in WordNet. Since highly ranked senses have more influence on the concept type score, the weighted scoring minimizes tilting of the final concept's type towards the rare meanings of its mentions.

An additional improvement to the preprocessing proposed by Hamborg et al. \cite{Hamborg2019a} is CoreNLP's coreference resolution on (virtually) concatenated documents. We observe that CoreNLP resolves more true mentions in concatenated texts than within single texts but is also prone to wrongly merging large coreferential chains, even when using the improved CR model \cite{clark-manning-2016-improving}. Thus, we split chains by analyzing NEL results of each of their mentions, i.e., for each mention and its compound+head sub-phrases, we obtain its Wikipedia page name \cite{wiki:2020}. We split chains whose members are assigned to different Wikipages \cite{minard-etal-2016-meantime}. We attach a resolved Wikipage title to each cleared chain as a new property. 

Each mention contains dependency and structure parsing subtrees. Structure subtrees are created by 1) parsing the sentences of mentions' origin, 2) mapping heads of mentions to the structure parsing trees, 3) taking the longest subtree but not larger than 20 tokens. The dependency subtree contains all the tokens as in the structure subtree. These subtrees play a role of the feature sources for the sieves.

\subsection*{S1: named entity linking (NEL)}
The first sieve, S1, leverages CoreNLP to mainly resolve mentions NE-containing chains, i.e., pronominal, nominal, and pronoun mentions. Specifically, S1 reused CoreNLP's coreference resolution on the concatenated documents and the assigned Wikipages. We resolve mentions by the winner-takes-it principle \cite{Hou2018a}, i.e., we merge the smaller chains into the coreference chains that were pre-identified by CoreNLP if the chains have identical Wikipage titles.

\subsection*{S2: head-word and compound match of NEs}
The second sieve, S2, merges the NE-containing chains, which mentions CoreNLP and NEL (in S1) failed to resolve. These typically smaller NE-containing chains of comparable types by $cm_2$ are merged into the larger NE-containing chains if they have identical NE heads (``\textit{Kim}'' -- ``North Korean dictator \textit{Kim} Jong Un'') or have cross-overlapping NE heads and compounds (``\textit{Donald}'' -- ``\textit{Donald} Trump'').

\subsection*{S3: resolution of non-NE mentions}
The third sieve, S3, merges any comparable NE-containing chain $c_{ne}$ with a smaller chain with non-NE mentions $c_{nn}$ if they match due to string or cosine similarity, e.g., ``Teresa May'' -- ``the prime minister.'' S3 focuses on using dependency subtrees of the mentions and extracts head modifiers from the subtrees, e.g., adjective, compound, apposition, or noun. Such anaphora tend to be missed by CoreNLP. In S3, we merge chains if at least one of the conditions holds: 
\newline 
1) $r(c_{nn}) \subset r(c_{ne}) \land  m(c_{nn}) \subset |r(c_{nn}) \cap r(c_{ne})|$, where $r(..)$ extracts representative phrases, i.e, phrases that contain heads of phrases and their direct modifiers (i.e., adjectival, noun, and compound) expanded with a list of all apposition modifiers, and $m(..)$ are modifiers (i.e., compounds and appositions) extracted from the chains' mentions,
\newline 
2) $|c_{nn} \cap c_{ne}| \geq 2$ and at least one head of phrases from each chain belongs to this intersection, 
\newline 
3) $cos(v(c_{ne}), v(c_{nn})) \geq t_{nn}$, where $v(..)$ is a mean vector of the word vector representation of all unique non-stop-words from the concept's mentions, $cos$ is cosine similarity and $t_{nn}$ is a threshold derived during experiments.

\subsection*{S4: identification of groups of persons}
The fourth sieve, S4, resolves chains of mentions referring to the group of individuals, e.g., ``illegal aliens'' -- ``undocumented immigrants,'' and semantically related phrases to countries, e.g., ``Trump administration officials'' -- ``American government.'' 

We incorporate a clustering approach of Zhukova et al. \cite{zhukova2021concept} that clusters mentions into chains in the decreasing semantic similarity between mentions. First, it finds distinctive cluster cores, i.e., each core is a group of mentions that are highly semantically similar to each other as to two factors: 1) cosine similarity of word-vectorized mentions in a potential core and 2) the normalized number of other mentions to which mentions of a potential core are similar. Then, the approach tries to assign the remaining mentions to the cores based on the relaxed rules of one-level similarity, i.e., using only cosine similarity between unresolved mentions and the already clustered mentions into chains. Mentions are to be left unresolved and converted into singleton-concepts if they do not meet similarity requirements, such as similarity to one of the core points. 

Lastly, because the annotation of GPEs include also mentions to the nations of the countries, we merge the country-NE-containing chains $c_{ne}$ with the groups of people $c_{gr}$ if $cos(v(h(c_{ne}),v(m(c_{gr})) \geq t_{gr}$, where $h(..)$ extracts heads of phrases, $m(..)$ extracts modifiers (i.e., NE-compounds and NE-adjectival)  from chains' mentions, and $t_{gr}$ is a threshold derived during experiments.

\subsection*{S5: events and abstract entities}
The fifth sieve, S5, resolves mention chains of events and so-called abstract entities \cite{poesio-artstein-2008-anaphoric}, i.e, actions, objects, events, etc, that contain mentions of noun phrases (NPs) and verb phrases (VPs), for example, ``Trump-Kim \textit{meeting}'' -- ``\textit{discussed} an issue''. Such mentions are typically not resolved by CoreNLP and contain either one the mentions or identical mentions. S5, first, keeps the unique mentions within chains, removes articles, and lemmatizes words. Second, S5 adapts a weighting scheme of word embedding vectors from Zhukova et al. \cite{zhukova2021concept}. S5 weights heads of mentions with a coefficient $k=2$ and keeps the original vectors for the other words in a mention. Such a weighting scheme allows emphasising the core meaning of a mention while keeping mention's context. Lastly, we form coreferential chains by clustering the mentions with hierarchical clustering \cite{murtagh2012algorithms} using cosine distance, average linkage, and a threshold $t_{cl}$. Hierarchical clustering commonly used in concept identification \cite{cambria2016senticnet}.

\section{Evaluation}
\label{sec:eval}
We compare \methodname/ to the established CDCR model of Barhom et al. \cite{barhom-etal-2019-revisiting} on NewsWCL50, i.e., a concept-driven CDCR dataset with a mix of identify and bridging relations. We evaluate the datasets containing coreference anaphora with bridging relations with the standard coreference metrics of CoNLL \cite{pradhan-etal-2012-conll}, although most of the bridging relations are not classified as a problem of set identification \cite{Hou2018a}.

\subsection{Dataset: NewsWCL50}
Similar to \cite{Cattan2021}, we do not evaluate the approaches on two separate lists of mentions for events and entities, but consider the mentions of all mentions combined as if they formed abstract entities. If an approach requires input or separate event and entity mentions (e.g., Barhom et al. \cite{barhom-etal-2019-revisiting}), then we implement splitting of mentions into two lists based on the heads of phrases, i.e., VPs represent events and other part-of-speech tags represent entities.

We removed ``ACTOR-I'' category from NewsWCL50 due to a large level of abstractnesses \cite{Hamborg2019a}. Our manual inspection of the chains labeled with this category did not identify consistency in annotating mentions as coreferential by identity, near-identity, or bridging relations. Therefore, to focus only on the listed relations, we removed this category from the evaluation.

\subsection{Methods and baselines}
\label{sec:methods}
We compare the performance of \methodname/ to a lemma baseline, an established event-entity CDCR model of Barhom et al. \cite{barhom-etal-2019-revisiting}, and Target Concept Analysis method (TCA) of Hamborg et al. \cite{Hamborg2019a} to identify semantic concepts in NewsWCL50. Additionally, we perform an ablation study and evaluate modifications of \methodname/. For all approaches, we use the same default parameters across topics, datasets, and run configurations to facilitate fair evaluation. We describe each method briefly in the following.

\subsubsection{Lemma} Our baseline is a primitive CDCR approach that resolves mentions based on matching lemmas of the phrases' heads. This baseline was also used by Barhom et al. \cite{barhom-etal-2019-revisiting} and establishes a fair comparison to the other approaches in the evaluation. 

\subsubsection{EeCDCR} Barhom et al. \cite{barhom-etal-2019-revisiting} proposed a joint event and entity CDCR model (hereafter EeCDCR). EeCDCR is trained on ECB+ and resolves event and entity mentions jointly. To reproduce EeCDCR's performance, we use the model's full set of optional features: semantic role labeling (SRL), ELMo word embeddings \cite{peters-etal-2018-deep}, and dependency relations. Barhom et al. \cite{barhom-etal-2019-revisiting} used the output files of the SRL parser, SwiRL, which makes it impossible to apply EeCDRCR to the other datasets. Therefore, we used the most recent AllenNLP's SRL method \cite{shi2019simple} to make the feature extraction a part of the EeCDCR and applicable for all datasets. To resolve intra-document mentions, identical to the original setup, we use Stanford CoreNLP. We reused default parameters for the model inference.

\subsubsection{TCA} Target Concept Analysis (TCA) is a method of identification of the  reported concepts in the related news articles, the mentions of which are typically a subject of bias by word choice and labeling \cite{Hamborg2019a}. Based on the functionality of TCA, we can classify the method as CDCR focusing on the resolution of concepts with varying strength of cross-mention identity relations and mentions of high lexical diversity. 

TCA uses six sieves to determine whether two candidate chains should be merged because they refer to the same semantic concept. Each sieve uses specific similarity measures, e.g., cosine similarity or string equality, and analyzes specific characteristics of the candidates' mentions, e.g., heads and their modifiers, i.e., adjective, noun, apposition, compound, and number. We use the reported TCA's default parameters for all datasets. For a word embedding model, we used a version of word2vec \cite{mikolov2013distributed} that unlike the original implementation vectorizes out-of-vocabulary words \cite{patel2018magnitude}.

Identical to Hamborg et al., we use TCA's default parameters for all datasets. We report the results of multiple variants of TCA. First, the original version as reported by Hamborg et al. Second, to facilitate comparability of TCA's sieves to the sieves of \methodname/, we use the preprocessing steps of \methodname/ in TCA and also show the effectiveness of these steps to improve the performance. Additionally, we evaluate TCA\textsubscript{preproc} using three different word embeddings (see below).

\subsubsection{Ablation study} For the ablation study, we test two variants of \methodname/ with which we investigate the effectiveness of various word embedding models and the effectiveness of approaches identifying more loose anaphoric coreference relations. 

First, an ``intermediate'' model named \methodname/\textsubscript{interm} uses sieves S1-S3 of \methodname/, S4\textsubscript{interm} is a baseline used by Zhukova et al. \cite{zhukova2021concept} with the same threshold parameters, and S5\textsubscript{interm} is the second sieve adopted from TCA, i.e., cosine similarity of phrases' heads. Resolution with semantically similar heads can effectively resolve mentions of abstract entities, e.g., ``meeting'' -- ``talks.'' S4\textsubscript{interm} and S5\textsubscript{interm} resolve mentions of the same concept types as in \methodname/. Using \methodname/\textsubscript{interm}, we test if the proposed methods for resolution of the bridging coreference anaphora outperform the simpler methods for the same coreference relations. 

Second, we test for either \methodname/ and \methodname/\textsubscript{interm} how using state-of-the-art, non-contextualized word embeddings affects their performance: word2vec \cite{mikolov2013distributed}, fastText \cite{mikolov2018advances}, and GloVe \cite{pennington-etal-2014-glove}. We use the model implementations that facilitate the representation of out-of-vocabulary (OOV) words \cite{patel2018magnitude}, which is critical to address the inability of the default word2vec and GloVE models to represent OOV words.

\subsection{Metrics}
\label{sec:metrics}
We report the established CoNLL metrics for (CD)CR, i.e., $MUC$, $B3$, $CEAF_e$, and an average of them as $F1_{CoNLL}$ \cite{pradhan-etal-2012-conll}. Similar to Barhom et al. \cite{barhom-etal-2019-revisiting}, we evaluate the methods with an official CoNLL scorer \cite{pradhan-etal-2014-scoring}. We do not distinguish between the strength of coreferential relations and evaluate the coreferential chains as if their mentions had relations of identical strength.

\begin{table}[]
\centering
\begin{tabular}{l|c| ccc|ccc|ccc|c}
\hline
\multirow{2}{*}{\textbf{Method}}         & \multirow{2}{*}{\textbf{Word vectors}}            & \multicolumn{3}{c|}{\textbf{MUC}} & \multicolumn{3}{c|}{\textbf{B3}} & \multicolumn{3}{c|}{\textbf{CEAF$_e$}} & \multirow{2}{*}{\textbf{F1$_{CoNLL}$}} \\
\cline{3-11}
           &                 & R      & P      & F1    & R      & P     & F1    & R      & P      & F1     &           \\
\hline
Lemma                   & --- & 75.7   & 93.8   & 83.8    & 36.8  & 88.8  & 52.0   & 63.1       & 7.8        & 13.9        & 49.9      \\
\hline
EeCDCR                  & GloVe           & 69.4   & 90.6   & 78.6    & 33.1  & 82.3  & 47.2   & 58.6       & 7.8        & 13.7        & 46.5      \\
\hline
TCA             & word2vec       & 73.4   & 89.4   & 80.6    & 37.2  & 73.7  & 49.4   & 51.9       & 8.7        & 14.9        & 48.3      \\
\hline
\multirow{3}{*}{TCA\textsubscript{preproc} }  & word2vec       & 72.9   & 89.5   & 80.3    & 38.4  & 75.6  & 50.9   & 54.5       & 9.1        & 15.6        & 48.9      \\
                        & fastText       & 72.9   & 87.6   & 79.6    & 37.3  & 71.5  & 49.0   & 52.0       & 9.5        & 16.0        & 48.2      \\
                        & GloVe         & 77.2   & 88.3   & 82.4    & 41.8  & 67.0  & 51.4   & 52.9       & 12.1       & 19.6        & 51.2      \\
\hline
\multirow{3}{*}{\methodname/\textsubscript{interm}}  & word2vec       & 68.4   & 90.3   & 77.8    & 37.7  & 84.0  & 52.0   & 63.0       & 8.4        & 14.8        & 48.2      \\
                        & fastText      & 74.2   & 87.3   & 80.2    & 38.7  & 71.5  & 50.2   & 58.4       & 11.6       & 19.4        & 50.0      \\
                        & GloVe          & 75.7   & 88.5   & 81.6    & 40.6  & 72.1  & 52.0   & 58.9       & 12.1       & 20.0        & 51.2      \\
\hline
\multirow{3}{*}{\methodname/} & word2vec       & 70.7   & 89.8   & 79.1    & 36.3  & 82.4  & 50.4   & 63.0       & 9.4        & 16.3        & 48.6      \\
                        & fastText       & 78.6   & 90.0   & 83.9    & 43.1  & 70.5  & 53.5   & 60.4       & 13.7       & 22.4        & 53.3      \\
                        & GloVe         & 79.3   & 90.8   & 84.7    & 44.4  & 72.2  & 55.0   & 61.1       & 13.9       & 22.6        & \textbf{54.1}  \\
\hline
\end{tabular}
\label{tab:eval}
\caption{Evaluation of a lemma baseline, EeCDCR \cite{barhom-etal-2019-revisiting}, TCA  \cite{Hamborg2019a}, \methodname/$_{interm}$ (a version of \methodname/ with baseline methods for sieves S4 and S5), and \methodname/ on the two diverse CDCR datasets: ECB+ and NewsWCL50.}
\end{table}

\begin{table}[]
\centering
\begin{tabular}{c| ccc|ccc|ccc|c}
\hline
\multirow{2}{*}{\textbf{Sieves}}            & \multicolumn{3}{c|}{\textbf{MUC}} & \multicolumn{3}{c|}{\textbf{B3}} & \multicolumn{3}{c|}{\textbf{CEAF$_e$}} & \multirow{2}{*}{\textbf{F1$_{CoNLL}$}} \\
\cline{2-10}
                         & R      & P      & F1    & R      & P     & F1    & R      & P      & F1     &           \\
\hline
init\textsubscript{shared}  & 28.6   & 89.8   & 43.4    & 15.6  & 96.0  & 26.8   & 41.4       & 2.1        & 4.1         & 24.7      \\
                                                                             S1\textsubscript{shared} & 40.2   & 92.3   & 56.0    & 22.9  & 95.0  & 36.9   & 49.7       & 3.1        & 5.8         & 32.9      \\
                                                                            S2\textsubscript{shared} & 42.2   & 92.5   & 58.0    & 24.6  & 94.6  & 39.1   & 51.2       & 3.3        & 6.2         & 34.4      \\
                                                                             S3\textsubscript{shared} & 45.7   & 91.0   & 60.8    & 27.3  & 91.3  & 42.1   & 51.6       & 3.6        & 6.7         & 36.5      \\
\hline
S4\textsubscript{interm} & 52.7   & 90.2   & 66.6    & 29.1  & 87.5  & 43.6   & 51.5       & 4.2        & 7.8         & 39.3      \\
                                         S5\textsubscript{interm} & 75.7   & 88.5   & 81.6    & 40.6  & 72.1  & 52.0   & 58.9       & 12.1       & 20.0        & 51.2      \\
\hline
S4 & 54.6   & 91.1   & 68.3    & 30.4  & 86.1  & 44.9   & 51.9       & 4.4        & 8.1         & \textbf{40.4}      \\
                                                                            S5 & 79.3   & 90.8   & 84.7    & 44.4  & 72.2  & 55.0   & 61.3       & 13.9       & 22.7        & \textbf{54.1}     \\
\end{tabular}
\label{tab:sieves}
\caption{Comparison of \methodname/'s sieves to the intermediate version of \methodname/$_{interm}$ in NewsWCL50 with GloVe word vectors. \methodname/ and \methodname/\textsubscript{interm} share sieves S1-S3 and differ in the last sieves. S4 and S5 of \methodname/ outperform sieves of \methodname/$_{interm}$.}
\end{table}

\begin{table}[]
\centering
\begin{tabular}{l|c| ccc|ccc|ccc|c}
\hline
\multirow{2}{*}{\textbf{Method}}         & \multirow{2}{*}{\textbf{Word vectors}}            & \multicolumn{3}{c|}{\textbf{MUC}} & \multicolumn{3}{c|}{\textbf{B3}} & \multicolumn{3}{c|}{\textbf{CEAF$_e$}} & \multirow{2}{*}{\textbf{F1$_{CoNLL}$}} \\
\cline{3-11}
           &                 & R      & P      & F1    & R      & P     & F1    & R      & P      & F1     &           \\
\hline
Lemma    & ---                                                       & 74.8   & 71.7   & 73.2    & 17.4  & 68.8  & 27.8   & 53.8       & 5.3        & 9.7         & 36.9      \\
\hline
EeCDCR & GloVe &	67.2 &	54.9 &	60.4	& 16.7	& 56.0 &	25.7 &	49.2 &	4.4 & 	8.0 & 	31.4 \\
\hline
TCA     &              word2vec                                     & 74.7   & 48.4   & 58.7    & 21.2  & 39.6  & 27.6   & 40.9       & 5.4        & 9.5         & 32.0      \\
\hline
\multirow{3}{*}{S3\textsubscript{interm}} &  word2vec  & 60.0   & 75.5   & 66.9    & 13.8  & 75.0  & 23.3   & 46.1       & 3.5        & 6.5         & 32.2      \\
 & fastText  & 71.6   & 68.7   & 70.1    & 21.8  & 60.7  & 32.1   & 38.8       & 4.4        & 8.0         & 36.7      \\
 & GloVe     & 72.6   & 72.1   & 72.3    & 19.8  & 63.0  & 30.1   & 38.1       & 4.5        & 8.1         & 36.8     \\
\hline
\multirow{3}{*}{S3}  &  word2vec & 80.0   & 69.2   & 74.2    & 27.3  & 55.0  & 36.5   & 45.7       & 7.2        & 12.5        & \textbf{41.1}      \\
& fastText & 81.5   & 60.1   & 69.2    & 33.2  & 42.7  & 37.4   & 34.1       & 6.1        & 10.3        & 39.0      \\
& GloVe    & 81.7   & 63.3   & 71.3    & 28.4  & 49.7  & 36.2   & 38.0       & 6.7        & 11.4        & \textit{\textbf{39.6} }     \\
\hline
\end{tabular}
\label{tab:s3}
\caption{Comparison of S4 for resolution groups of persons between \methodname/ to the intermediate version of \methodname/$_{interm}$.}
\end{table}

\begin{table}[]
\centering
\begin{tabular}{p{0.15\textwidth}|p{0.85\textwidth}}
\hline
\makecell[c]{\textbf{Name}} & \makecell[c]{\textbf{Resolved mentions}} \\
\hline
PRK-USA Summit & the summit meeting, a potential meeting of the two leaders,
an extraordinary meeting following months of heated rhetoric, 
meet with the North Korean dictator, 
discuss its nuclear weapons program, 
Kim's offer for a summit, 
a great chance to solve a world problem, 
won't even have a meeting at all, 
a once-unthinkable encounter between him and Mr. Kim, 
a one-on-one meeting with North Korea leader Kim Jong Un, 
direct talks between U.S.President Donald Trump and Kim, 
Mr. Kim's invitation to meet, 
the upcoming summit meeting with the North Korean leader, 
a great chance to solve a world problem,
unwavering determination in addressing the challenge of North Korea \\
\hline
DNC's lawsuit & the process of legal discovery, 
a sham lawsuit about a bogus Russian collusion claim, a bogus Russian collusion claim, 
allegations of obstruction of justice, 
a desperate attempt to keep a collusion narrative going ahead of November mid-term elections,  a new low to raise money, 
the DNC's move, 
the lawsuit to drum up donations for the party \\
\hline
Denucleari- zation & a deal to destroy only inter-continental missiles that could reach the United States, 
gives up nuclear weapons,  months of heated rhetoric over Pyongyang's nuclear weapons program, 
to engage in a process headed toward an ambiguous goal, 
broad and "abstract " statements about the need for North Korea to ``denuclearize'', 
give up its nuclear program, 
yet to take any tangible steps to give up its nuclear arsenal,          
to address the threats posed by its nuclear and missile program \\
\hline
Coming into the US & made their way to the U.S.-Mexico border, 
begun crossing into the U.S.,
the arrival of a caravan of Central American migrants,
the arrival of a few hundred, 
crossing through a legal port of entry, 
driving families to flee,
the caravan's steady approach to the U.S.,
to pass through into our country,
several attempted illegal entries by people associated with the caravan,
may be detained or fitted with ankle monitors and released,
wait to be processed by U.S. authorities,
to turn themselves in,
a process that unfolds over several months or longer,
made their way toward the border,
a test of President Trump's anti-immigrant politics,
a volatile flash point in the immigration debate ignited by Mr.Trump,
headed north together as a form of protection,
the final act of the caravan
 \\
\hline
Immigrants & no legitimate asylum-seekers, 
the asylum-seekers, 
the individual migrants planning to seek asylum, 
groups of the migrants with their children, 
migrant families that request asylum, 
unauthorized immigrants, 
the migrants in the caravan, 
a caravan of immigrants, 
members of the caravan, 
these large ``Caravans'' of people, 
a few hundred asylum seekers, 
refugees, 
those individuals, 
applicants, 
people who request protection, 
people traveling without documents, 
several groups of people associated with the caravan, 
undocumented immigrants, 
asylum-seeking immigrant ``caravan'',
group of about 100 people
\\
\hline
\end{tabular}
\label{tab:examples}
\caption{Examples of mentions resolved by \methodname/. The table shows the annotated concept names and only unique resolved mentions.}
\end{table}

\subsection{Results and Discussion}
\label{sec:res}

The evaluation shows that \methodname/, our multi-sieve feature-based unsupervised method, is capable of consistently resolving coreferential anaphora of mixed identity and bridging coreference relations with $F1_{CoNLL}=54.1$ when using GloVe as word vectors.  

\Cref{tab:eval} reports the $F1_{CoNLL}$ score typically used to evaluate (CD)CR approaches (named $F1$ in this section). We find that \methodname/ outperforms all methods on NewsWCL50 dataset ($F1 = 54.1$). \methodname/ performs better than a simple lemma-baseline by $F1_{\Delta} = 4.2$ and outperforms TCA, i.e., a previously used method for the concept identification in NewsWCL50 dataset, by $F1_{\Delta} = 5.8$. \methodname/ shows that the mentions of the mixed identity and bridging relations are resolved if treating all mentions without an upfront separation by the coreference strength of the anaphora. Moreover, learning context can substituted by feature engineering from the phrases' extracted nearest context, i.e., parsing subtrees. 

When comparing TCA to its modification TCA\textsubscript{preproc}, we see that the preprocessing steps employed in \methodname/ improve the overall performance, i.e., the improved identification of concept type and initial coreference resolution by CoreNLP on the combined documents instead on single documents. Using TCA with GloVe wordvectors improves the performance of original TCA by $F1_{\Delta} = 2.9$. We assume that the concentration of the NE mentions in NewsWCL50 is high and such preprocessing step positively impacts the overall performance. 

\Cref{tab:sieves} compares the sieves of \methodname/ to the``intermediate'' version \methodname/\textsubscript{interm}. Sieves S4 and S5 outperform S4\textsubscript{interm} and S5\textsubscript{interm} by $\Delta_{S4} = 1.1$ and $\Delta_{S5} = 1.8$ correspondingly. Additionally, we evaluate S4 on the mentions annotated as groups of persons. \Cref{tab:s3} shows that S4 in the original implementation by Zhukova et al. \cite{zhukova2021concept} with word2vec as word embedding performs best and the implementation with GloVe is second best. We assume that the thresholds of the proposed approach are better suited for word2vec and could be tuned for GloVe. 

\Cref{tab:examples} shows multiple examples of how \methodname/ manages to resolve the annotated mentions that belong to the complex abstract entities. The table shows that some mentions form coreference relations only within a given context. For example, a mention ``a great chance to solve world problem,'' which describes the summit between Donald Trump and Kim Jong Un, forms a positive association ``the summit = a great chance'' via a context-specific bridging relation. A coreference link ``a one-to-one meeting with North Korean leader Kim Jong Un'' -- ``a once unthinkable encounter between Trump and Mr. Kim'' indicates similar positive bias via  the chosen word choice. 

NewsWCL50 dataset shows that CDCR datasets with a mix of identity and bridging, i.e., looser, coreference relations represent a challenge to the established models, such as EeCDCR (performs worst among all baselines). A new direction in CDCR research proposed Caciularu et al. \cite{Caciularu2021} by training a cross-document language model to enable CDCR models to understand the broad and narrow context of the mentions. Such deep context learning will improve the representation of phrases' semantics and cross-phrase relations.

\section{Conclusion}

We propose \methodname/, an unsupervised sieve-based method for cross-document coreference resolution (CDCR) that unlike the state-of-the-art CDCR methods resolves mentions of a mix of strict and loose coreference relations, e.g., ``American steelmakers'' -- ``shuttered plants and mills,'' and ``the United States'' - ``Trump Administration officials.'' \methodname/ performs best among the evaluated approaches and also outperforms a previous approach for resolution of such complex coreferential chains by $F1_{CoNLL}=5.8$. Further, a well-established CDCR model performs worse on NewsWCL50. Our findings suggest that CDCR models need to be tested on more diverse CDCR datasets that contain both strict identity and more loose bridging coreference relations. In political news articles, i.e., a challenging ``wild'' environment for the application of CDCR approaches, such relations might create the biased perceptions of reported entities and concepts. Therefore, resolution of mentions with context-specific coreference relations is a step towards bringing awareness of bias by word choice and labeling and increase the level of complexity of CDCR research.

\bibliography{custom}
\bibliographystyle{splncs04}

\end{document}